\documentclass[10pt,twocolumn,letterpaper]{article}

\usepackage{iccv}
\usepackage{times}
\usepackage{epsfig}
\usepackage{graphicx}
\usepackage{amsmath}
\usepackage{amssymb}
\usepackage{color}
\usepackage{caption}
\usepackage{booktabs}
\usepackage[numbers,sort]{natbib}
\usepackage{multirow}
\usepackage{subcaption}

\definecolor{hollywoodcerise}{rgb}{0.96, 0.0, 0.63}
\definecolor{lasallegreen}{rgb}{0.03, 0.47, 0.19}
\definecolor{hanpurple}{rgb}{0.32, 0.09, 0.98}
\definecolor{green(pigment)}{rgb}{0.0, 0.65, 0.31}

\usepackage[pagebackref=true,breaklinks=true,letterpaper=true,colorlinks,bookmarks=false]{hyperref}

\hypersetup{colorlinks,linkcolor={red},citecolor={hollywoodcerise},urlcolor={black}}

\iccvfinalcopy 

\def\iccvPaperID{1576} 

\ificcvfinal\fi

\begin{document}

\title{Dual Transfer Learning for Event-based End-task Prediction via Pluggable Event to Image Translation}

\author{Lin Wang, Yujeong Chae, and Kuk-Jin Yoon\\
Visual Intelligence Lab., KAIST, Korea\\
{\tt\small \{wanglin, yujeong, kjyoon\}@kaist.ac.kr}
}

\maketitle
\ificcvfinal\thispagestyle{empty}\fi

\begin{abstract}
  Event cameras are novel sensors that perceive the per-pixel intensity changes and output asynchronous event streams with high dynamic range and less motion blur. 
 It has been shown that events alone can be used for end-task learning, \eg, semantic segmentation, based on encoder-decoder-like networks. However, as events are sparse and mostly reflect edge information, it is difficult to recover original details merely relying on the decoder. Moreover, most methods
 resort to the pixel-wise loss alone for supervision, which might be insufficient to fully exploit the visual details from sparse events, thus leading to less optimal performance.
  In this paper, we propose a concise yet effective dual-branch framework named Dual Transfer Learning (DTL) to enhance the performance on the event-based end-task prediction.
  The proposed approach is composed of three modules: event to end-task learning (EEL) branch, event to image translation (EIT) branch, and transfer learning (TL) module that simultaneously explores the feature-level and pixel-level knowledge from the EIT branch to improve the EEL branch. This simple yet novel method leads to strong representation learning from events and is evidenced by the significant performance boost on the end-tasks such as semantic segmentation and depth estimation. 
\end{abstract}

\vspace{-12pt}
\section{Introduction}
\begin{figure}[t!]
    \centering
    \captionsetup{font=small}
    \includegraphics[width=.98\columnwidth]{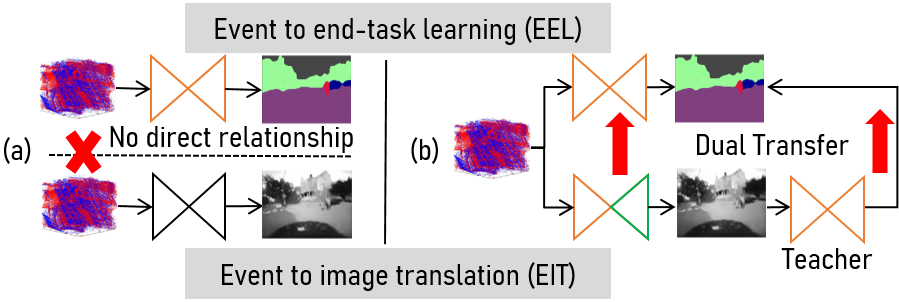}
    \vspace{-9pt}
    \caption{(a) There is no direct relation between EIT and EEL in the prior-arts. (b) The proposed DTL framework by using EIT branch as a pluggable unit and transferring both feature-level and prediction-level information to enhance the performance of EEL.}
    \label{fig:cover}
\vspace{-12pt}
\end{figure}

\begin{figure}[t!]
    \centering
    \captionsetup{font=small}
    \includegraphics[width=.94\columnwidth, height=3cm]{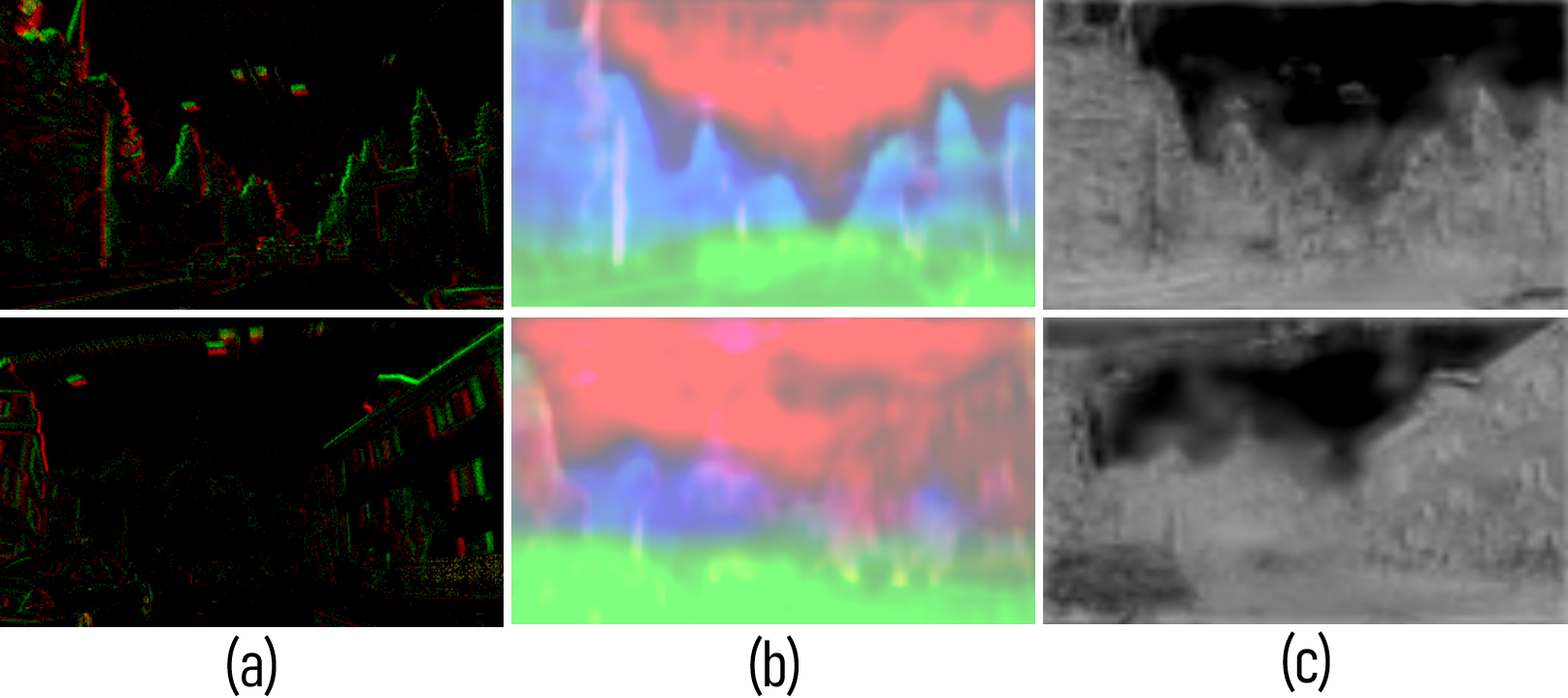}
    \vspace{-8pt}
    \caption{Visualization of features from EEL and EIT decoders of the same input. (a) Events, (b) EEL features, (c) EIT features.}
    \label{fig:fea_vis}
    \vspace{-15pt}
\end{figure}

\vspace{-3pt}
Event cameras have recently received much attention in the computer vision and robotics community for their distinctive advantages, \eg, high dynamic range (HDR) and less motion blur~\cite{Gallego2020EventbasedVA}. 
These sensors perceive the intensity changes at each pixel asynchronously 
and produce event streams encoding time, pixel location, and polarity (sign) of intensity changes.
Although events are sparse and mostly respond to the edges in the scene, it has been shown that it is possible to use events alone for learning the end-tasks, \eg, semantic segmentation~\cite{alonso2019ev, gehrig2020video}, optical flow and depth estimation~\cite{tulyakov2019learning,hidalgo2020learning, zhu2019unsupervised,gehrig2019end}, via deep neural networks (DNNs). 
These methods usually follow the encoder-decoder-like network structures, \eg, \cite{alonso2019ev,hidalgo2020learning,gehrig2020video}, as shown in Fig.~\ref{fig:cover}(a), and are trained in a fully supervised manner. 
However, as events mostly reflect edge information, unlike the canonical images, it is difficult to recover the original structural details from events merely relying on the decoder (see Fig.~\ref{fig:fea_vis}(b)). Importantly, learning from sparse events with the pixel-wise loss (\eg, cross-entropy loss) alone for supervision often fails to fully exploit visual details from events, thus leading to less optimal performance. 

The other line of research has shown the possibility of generating images from events
\cite{rebecq2019high, wang2020event, wang2019event, stoffregen2020reducing,hu2020learning, wang2020joint, mostafavilearning}.
The generated images have been successfully applied to the end-tasks, \eg, object recognition \cite{rebecq2019high}; however, there exist two crucial problems. First, using these images as the intermediate representations of events
leads to considerable inference latency. Second, there is no direct  connection regarding the optimization process between two tasks.
Indeed, the feature representations learned from image generation contain more structural details (see Fig.~\ref{fig:fea_vis}(c)), which can be a good guide for learning end-tasks. However, these crucial clues have been rarely considered and explored to date. Moreover, some event cameras provide active pixel sensor (APS) frames, which contain very crucial visual information; nonetheless, the potential has been 
scarcely explored to assist learning end-tasks from events. 

\begin{figure*}[t!]
    \centering
    \captionsetup{font=small}
    \includegraphics[width=.97\textwidth, height=5.2cm]{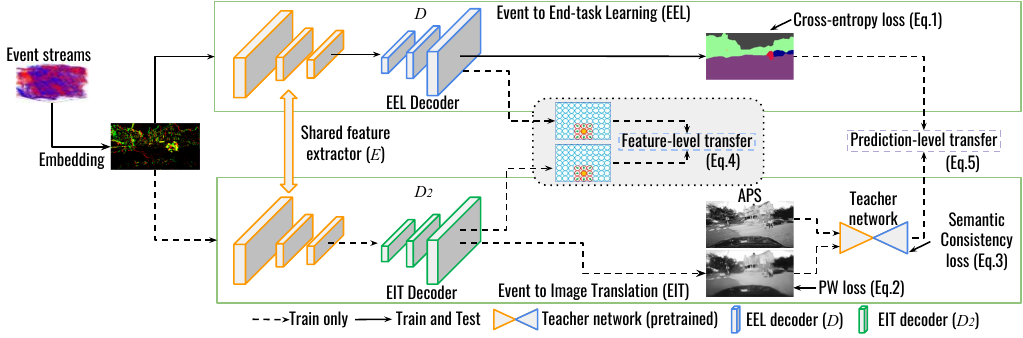}
    \vspace{-7pt}
    \caption{Overview of the proposed DTL framework, which consists of three parts: Event to End-task Learning (EEL) branch, Event to Image Translation (EIT) branch and Transfer Learning (TL) module. The dash line indicates `for training only'. }
    \label{fig:method}
    \vspace{-10pt}
\end{figure*}

Therefore, in this paper, we design a concise yet flexible framework to alleviate the above dilemmas. Motivated by recent attempts for event-to-image translation \cite{rebecq2019high, wang2020event, wang2019event,paredes2020back}, transfer learning \cite{zhuang2020comprehensive,liu2019structured,wang2020dual,deng2019towards,mou2020plugnet}, and multi-task learning \cite{vandenhende2020multi},   
we propose a novel Dual Transfer Learning (DTL) paradigm to efficiently enhance the performance of the end-task learning (see Fig.~\ref{fig:method}). Our DTL method follows a dual-branch pipeline, which
consists of three components: Event to End-task Learning (EEL) branch, Event to Image Translation (EIT) branch and Transfer Learning (TL) module, as shown in Fig.~\ref{fig:cover}(b). Specifically, we integrate the idea of image translation to the process of end-task learning, thus formulating the EIT branch. The EEL branch is then significantly enhanced by the TL module, which exploits to transfer the feature-level and prediction-level information from EIT branch. 
In particular, a affinity graph transfer loss~\cite{liu2019structured,turaga2009maximin,briggman2009maximin} is leveraged to maximize the feature-level instance similarity along the spatial locations between EEL and EIT branches. The prediction-level information is transferred from the EIT branch to the EEL branch using the APS and translated images based on a teacher network trained using the canonical images on the end-task.
Moreover, we share the same feature encoder between EEL and EIT branches and optimize them in an end-to-end manner. We minimize the prediction gap between the APS and translated images based on the teacher network and subtly leverage the supervision signal of the EEL branch to enforce semantic consistency for the EIT branch, which surprisingly helps to recover more semantic details for image translation. \textit{Once training is done, the EIT branch and TL module can be freely removed, adding no extra inference cost}.

We conduct extensive experiments on two end-tasks, semantic segmentation (Sec.~\ref{seg_exp}) and depth estimation (Sec.~\ref{depth_sec}). The results show that this simple yet novel method brings significant performance gains for both tasks. As a potential, our method can also learn the end-tasks via the teacher network \textit{without} using ground truth labels.  Although the EIT branch is regarded as an \textit{auxiliary} task, the results demonstrate that our DTL framework contributes to recover better semantic details for image translation.

In summary, our contributions are three folds. (I) We propose a novel yet concise DTL framework for the end-task learning. (II) We design a TL module where we transfer both feature-level and prediction-level information to the end-tasks.  (III) We conduct extensive experiments on two typical end-tasks, showing that DTL significantly improves the performances while adding no extra inference cost. We also demonstrate that DTL recovers better semantic details for the EIT branch. Our project code is available at \url{https://github.com/addisonwang2013/DTL}.

\vspace{-2pt}
\section{Related Works}
\vspace{-2pt}
\noindent \textbf{DNNs for event-based vision.} DNNs with event data was first explored in the classification \cite{neil2016phased} and robot control \cite{moeys2016steering}. \cite{maqueda2018event} then trained a DNN for steering angle prediction on the DDD17 dataset \cite{binas2017ddd17}. This dataset has been utilized in \cite{alonso2019ev,gehrig2020video,wang2021evdistill} to perform semantic segmentation using pseudo labels obtained from the APS frames. Moreover, DNNs have been applied to high-level tasks, such as 
object detection and tracking \cite{cannici2019asynchronous,hu2020learning,messikommer2020event,Chen_2020}, human pose estimation \cite{wang2019ev,xu2020eventcap,calabrese2019dhp19}, motion estimation \cite{stoffregen2019event,mitrokhin2020learning,kepple2020jointly,wang2020stereo}, 
object recognition \cite{gehrig2020video, rebecq2019high, bi2019graph} on N-Caltech \cite{orchard2015converting} and other benchmark datasets \cite{sironi2018hats,li2017cifar10,bi2019graph}.  

Meanwhile, another line of research focuses on the low-level vision tasks, such as optical flow estimation \cite{zhu2019unsupervised, gehrig2019end, stoffregen2020train,gallego2019focus}, depth estimation \cite{zhu2019unsupervised, tulyakov2019learning,gehrig2021combining, mostafavilearning} on the MVSEC dataset \cite{zhu2018multivehicle}. In addition, 
\cite{wang2019event, rebecq2019high,scheerlinck2020fast,stoffregen2020reducing} attempted to generate video from events using camera simulator \cite{rebecq2018esim, mueggler2017event}, and \cite{wang2020eventsr, mostafavi2020learning,wang2020event} tried to generate high-resolution images. 
In contrast,
\cite{gehrig2020video} proposed to generate events from video frames. Some other works explored the potential of events for image deblurring \cite{haoyu2020learning,jiang2020learning, wang2020eventsr}, HDR imaging \cite{han2020neuromorphic,zhang2020learning}, and event denoising \cite{baldwin2020event}.
 For more details about event-based vision, refer to a survey \cite{Gallego2020EventbasedVA}. Differently, we propose a DTL framework to enhance the performance for the end-task learning by exploring the knowledge from the EIT branch sharing the same encoder. We regard the EEL branch as main task and EIT branch as auxiliary one. As a potential, the EIT is also improved by the DTL framework. 
 
\vspace{2pt}
\noindent \textbf{Transfer Learning (TL).} TL aims to improve learning in a new task through the transfer of knowledge from a related task that has already been learned \cite{zhuang2020comprehensive}. Among the techniques, knowledge transfer (KT) is a typical approach for learning model with softmax labels or feature information of a learned model \cite{romero2014fitnets,hinton2015distilling,wang2020knowledge}. 
 Most KT methods
 focus on the image data and transfer knowledge using logits \cite{chen2019online,zhang2018deep,xu2020knowledge,yao2020knowledge} or features \cite{romero2014fitnets, kim2018paraphrasing, zagoruyko2016paying,park2019feed,heo2019comprehensive,wang2021semi}. Recently, some approaches have been proposed to transfer knowledge across different paired modality data with common labels \cite{gupta2016cross, hafner2018cross,zhao2020knowledge,yuan2019ckd,li2020towards,xu2018pad,pande2019adversarial,hu2020learning}. 
 For more details of TL, refer to recent surveys, \eg, \cite{wang2020knowledge,zhuang2020comprehensive}. Differently, we transfer knowledge across the EIT branch and the EEL branch. We do not explore multiple input modalities with multiple task networks. Instead, our input is homogeneous and tasks to be learned are not the same. Moreover, the networks share the same encoder, and the knowledge is transferred in two aspects: feature-level transfer via affinity graph learning and prediction-level transfer via a teacher network.

\vspace{2pt}
\noindent \textbf{Multi-task learning.} It aims to learn multiple tasks, \eg, object recognition, detection and semantic segmentation, jointly by leveraging the domain-specific information contained in the training signals of relevant tasks \cite{zhang2019pattern,xu2018pad,vandenhende2020mti,wang2021psat,wang2020dual,mou2020plugnet}. For more details, refer to a recent survey \cite{vandenhende2020multi}. These methods usually treat the multiple tasks equally in both training and inference. However, different from these methods for joint task learning in the same modality, we learn from cross-modalities and regard the EEL branch as the main and EIT branch as the auxiliary one.  The EIT branch and TL module can be flexibly removed during inference, adding no extra computation cost.

\vspace{-2pt}
\section{The Proposed Approach}
\vspace{-4pt}
\noindent \textbf{Event Representation.}
\label{event_rep}
 As DNNs are designed for image-/tensor-like inputs, we first describe the way of event embedding. An event $e$ is interpreted as a tuple $(\textbf{u}, t, p)$, where $\textbf{u}= (x,y)$ is the pixel coordinate, $t$ is the timestamp, and $p$ is the polarity indicating the sign of brightness change. An event occurs whenever a change in log-scale intensity exceeds a threshold $C$. 
A natural choice is to encode events in a spatial-temporal 3D volume to a voxel grid \cite{rebecq2019high,zhu2019unsupervised, zhu2018unsupervised} or event frame \cite{rebecq2017real,gehrig2019end} or multi-channel image \cite{wang2020event, lin2020learning,wang2019event}. In this work, 
we represent events to multi-channel images as the inputs to the DNNs, as done in \cite{wang2020event,wang2019event,lin2020learning}. More details about event representation are in the suppl. material.

\vspace{-3pt}
\subsection{Overview}
\vspace{-3pt}
For event cameras, \eg, DAVIS346C with APS frames, assume that we are given the  dataset $\mathcal{X} = \{e_i, {x_{aps}}_i, y_i\}$, where $e_i$ is $i$-th stacked multi-channel event image and $x_{{aps}_i}$ is the corresponding $i$-th APS image with its ground truth label $y_i$. 
Our goal is to learn an effective end-task model on the end-tasks, \eg, semantic segmentation, from the events. Existing methods \cite{alonso2019ev, gehrig2020video} rely on the encoder-decoder network structures and are trained using the ground truth (GT) labels for supervision via, \eg, cross-entropy loss. However, as events are sparse and mostly reflect the edges of the scene, it is difficult to recover original details merely relying on the decoder, as shown in Fig.~\ref{fig:fea_vis}(b). 
To address the dilemmas, we propose a novel yet effective framework, called dual transfer learning (DTL), to effectively learn the end-tasks. 
As shown in Fig.~\ref{fig:method}, the DTL framework consists of three components: (a) Event to End-task Learning (EEL) branch; (b) Event to Image Translation (EIT) branch, and Transfer Learning (TL) module.
The TL module transfers knowledge from the EIT branch to learn better representation of events for the EEL branch, without adding extra computation cost in the inference time.  We now describe these components in detail.

\vspace{-3pt}
\subsection{Dual Transfer Learning}
\vspace{-3pt}
\subsubsection{Event to End-task Learning (EEL)}
\vspace{-3pt}
For the end-tasks, \eg, semantic segmentation,  we simply adopt an encoder($E$)-decoder($D$) network \cite{chen2017rethinking}. The whole process of learning this branch is called Event to End-task Learning (EEL), as shown in ~Fig.~\ref{fig:method}. 
The EEL branch generates an output of prediction, \eg, label map, from an embedded event image in a dimension of W$\times$H$\times C$, where W, H and $C$ are the width, height and number of channels. This setting is similar to the methods for semantic segmentation \cite{alonso2019ev, gehrig2020video}, in which a conventional multi-class cross-entropy (CE) loss is used for supervision,  
which is defined as:
\begin{equation}
\setlength{\abovedisplayskip}{-1pt}
\label{ce_loss}
    \mathcal{L}_{CE} = \frac{1}{N}\sum_{i=1}^{N} -y_i\log(p_i)
\setlength{\belowdisplayskip}{-1pt} 
\end{equation}
where $N$ is the total pixel numbers, $p_i$ and $y_i$ refer to the predicted probability and GT label for pixel $i$. For the end-tasks, \eg, depth estimation, the loss for supervision can be flexibly changed to other pixel-wise loss, such as $L_1$ loss. 

However, we notice that using the supervision loss only is insufficient to fully exploit the visual information from events.  Furthermore, it is difficult to recover original details only relying on the decoder, as shown in Fig.~\ref{fig:fea_vis}(b).  The reason is that events respond predominantly to edges of the scenes, making the event data intrinsically sparse. This renders dense pixel-wise predictions from events challenging, especially in low contrast change or less motion regions.

To this end, we draw attention from recent attempts for events to image translation (EIT) \cite{wang2020event, rebecq2019high, hu2020learning}. Our motivations are two-folds. Firstly, given the same event input, we find that the feature representations of the EIT decoder contain more complete structural information of scenes, as shown in Fig.~\ref{fig:fea_vis}(c).  Secondly, the APS frames synchronized with event sensor and the translated images acting as intermediate representation of events can be fully leveraged 
for guiding the EEL branch.
To this end, we design an EIT branch and further exploit to transfer both feature-level and prediction-level knowledge to the EEL branch. 
We will discuss the details in the following Sec.~\ref{eit_sec}.

\begin{figure}[t!]
    \centering
     \captionsetup{font=small}
    \includegraphics[width=\columnwidth, height=3.0cm]{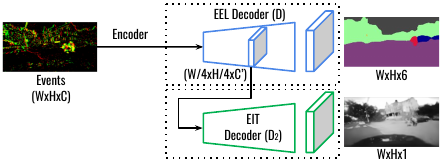}
    \vspace{-18pt}
    \caption{Designing the EIT decoder based on EEL branch.}
    \label{fig:dtl_fig}
    \vspace{-8pt}
\end{figure}

\vspace{-8
pt}
\subsubsection{Event to Image Translation (EIT)} 
\vspace{-3pt}
\label{eit_sec}
Under the similar network structure, the feature maps of EIT contain fine-grained visual structural information of scenes, as shown Fig.~\ref{fig:fea_vis}(c). Although these structural information does not convey the object class information, they can be effectively categorized and optimized between pixel to pixel or region to region.  That is, these categorized information indeed delivers crucial semantic knowledge, which can benefit the end-tasks. 

\begin{figure}[t!]
    \centering
     \captionsetup{font=small}
    \includegraphics[width=.75\columnwidth]{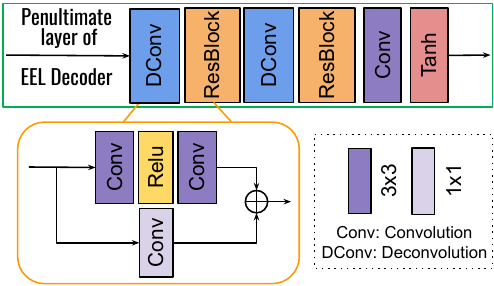}
    \vspace{-5pt}
    \caption{The proposed EIT decoder network structure.}
    \label{fig:eit_decoder}
    \vspace{-8pt}
\end{figure}

For the EIT branch, as the input is the same as the EEL branch, the encoded latent spaces from both branches are similar. Inspired by the design in~\cite{wang2020dual,liu2019structured,deng2019towards},
we share the same encoder $E$ for EEL and EIT branches to extract visual features, as shown in Fig.~\ref{fig:dtl_fig}. As the EEL decoder $D$ is not effective enough to reconstruct the fine-grained structure information of image, we design a decoder $D_2$ to generate high-quality results while reducing the computation. The detailed structure of $D_2$ is depicted in Fig.~\ref{fig:eit_decoder}. $D_2$ is based on the penultimate layer (with a dimension of 
W/4$\times$H/4$\times C'$) of EEL decoder $D$, which is further extended by designing ResBlocks connected by the deconvolution layers, followed by one 3x3 convolution (conv.) layer and a Tanh function. In particular, the ResBlock consists of a residual connection that takes one 3x3 conv. layer, followed by a ReLU function and one 3x3 conv. layer, helping to enlarge the receptive field. Interestingly, we find that adding Tanh function is crucial for image reconstruction. Based on the shared encoder $E$ and decoder $D_2$, we reconstruct images from events under the supervision of the APS frames using a pixel-wise loss (\eg, $L_1$ loss), which can be formulated as follows:
\vspace{-4pt}
\begin{equation}
\label{eit_loss}
    \mathcal{L}_{EIT} = \mathbb{E}_{e_i \sim \mathcal{X}}[||x_{{aps}_i} - D_2(E(e_i))||_1].
\end{equation}

To better preserve the semantic information, we exploit a novel semantic consistency (SC) loss based on a  teacher network $T$ pretrained on the end-task with canonical images, as shown in Fig.~\ref{fig:method}. The proposed SC loss for EIT has three advantages. Firstly, the generated image $D_2(E(e_i))$ becomes the optimal input of $T$. Secondly, the predictions from both the APS frames and generated images can provide auxiliary supervision for the EEL branch.
The EIT branch
can be removed after training, adding no extra inference cost.
The proposed SC loss is formulated as follows: 
\vspace{-5pt}
\begin{equation}
\setlength{\belowdisplayskip}{-0.2pt} 
\label{sc_loss}
 \mathcal{L}_{SC} = \mathbb{E}_{e_i \sim \mathcal{X}} KL[T(D_2(E(e_i)))||T(x_{aps_i})]  
 \setlength{\belowdisplayskip}{-0.2pt}
\end{equation}
where $KL$($\cdot||\cdot$) is the Kullback-Leibler divergence between two distributions.

\begin{table*}[t!] 
\renewcommand{\tabcolsep}{15pt}
 \centering
 \small
 \captionsetup{font=small}
 \caption{Segmentation performance with different event representations and APS frames on the test data \cite{alonso2019ev}, measured by Acc. (Accuracy) and MIoU (Mean Intersection over Union). The models are trained on time intervals of $50$ms but tested with $50$ms, $10$ms and $250$ms.}
 \vspace{-8pt}
 \begin{tabular}{l|c|c|c|c}
\hline
 Method & Event representation& MIoU [50ms] & MIoU [10ms] & MIoU [250ms]\\
\hline
EvSegNet~\cite{alonso2019ev} & 6-channel~\cite{alonso2019ev}& 54.81 & 45.85 &47.56 \\
Vid2E~\cite{gehrig2020video} & EST~\cite{gehrig2019end} &  45.48  &30.70  & 40.66\\
\hline
Ours-Deeplabv3 (Baseline) & Multi-channel &  50.92 & 41.61 & 43.87 \\
Ours-Deeplabv3 (DTL) &  Multi-channel &  \textbf{58.80} (+7.88) & \textbf{50.01} (+8.40) & \textbf{52.96} (+9.09) \\
\hline
 \end{tabular}
\label{tab:comp_table1}
\vspace{-8pt}
\end{table*}

\vspace{-5pt}
\subsubsection{Transfer Learning (TL) Module}
\label{kd_module}
\vspace{-3pt}
\noindent \textbf{Feature-level Transfer.} 
As the feature representations of the decoder $D_2$ for the EIT branch  (see Fig.~\ref{fig:fea_vis}(c)) deliver fine-grained visual structural information of scenes, we leverage these visual knowledge to guide the feature representation of the decoder $D$ of the EEL branch.  Motivated by~\cite{wang2020dual,liu2019structured, deng2019towards,hou2020inter}, we aim to
transfer the instance-level similarity along the spatial locations between EIT branch and EEL branch based on affinity graphs, as formulated in Eq.~\ref{affinity_graph}.
The node represents a spatial location of an instance (\eg, car), and the edge connected between two nodes represents the similarity of pixels. For events, if we denote the connection range (neighborhood size) as $\sigma$, then nearby events within $\sigma$ (9 nodes in Fig.~\ref{fig:method}) are considered for computing affinity contiguity. It is possible to adjust each node's granularity to control the size of the affinity graph; however, as events are sparse, we do not consider this factor. In such a way, we can aggregate top-$\sigma$ nodes according to the spatial distances and represent the affinity feature of a certain node. For a feature map $F \sim \mathbb{R}^{C\times H \times W}$ ($H \times W$ is the spatial resolution and $C$ is the number of channels), the affinity graph contains nodes with $H \times W \times \sigma$ connections. We denote $A_{ab}^{EIT}$ and $A_{ab}^{EEL}$ are the affinity graph between the $a$-th node and the $b$-th node obtained from the EIT  and EEL branch, respectively, which is formulated as~\cite{wang2020dual,liu2019structured, deng2019towards,hou2020inter}:
\vspace{-3pt}
\begin{equation}
\label{affinity_graph}
    \mathcal{L}_{FL} = \frac{1}{H\times W\times\sigma}\sum_{a\sim R}\sum_{b\sim \sigma}||A_{ab}^{{EIT}} -A_{ab}^{{EEL}}||_2^2 \\[-3pt]
\vspace{-1pt}
\end{equation}
where $R=\{1,2,\cdots, H \times W \}$ indicates all the nodes in the graph. 
The similarity between two nodes is calculated from the aggregated features $F_a$ and $F_b$ as: $A_{ab}= \frac{F_a^{\intercal} F_b}{||F_a||_2 || F_b||_2}$~\cite{turaga2009maximin,liu2019structured}, 
where $F_a^{\intercal}$ is the transposed feature vector of $F_a$. More details of the proposed feature-level (FL) transfer loss is provided in the suppl. material.

\vspace{2pt}
\noindent \textbf{Prediction-level Transfer.}
In addition to transferring the structural information of the feature representations from the EIT decoder $D_2$,  we observe that it is potential to leverage the paired APS frames to enhance EEL branch. 
In particular, inspired by the recent attempts for cross-modal learning \cite{gupta2016cross,zhao2020knowledge}, we aim to transfer the knowledge from the teacher network $T$ using the APS frames to the EEL network.
We view the segmentation problem as a collection
of separate pixel labeling problems, and directly strive to align the class probability of each pixel produced by the EEL network with that by the teacher network. We use the class probabilities produced from $T$ as soft targets for training the EEL network $E(D(\cdot))$. The prediction-level transfer loss is formulated as:
\vspace{-2pt}
\begin{equation}
\label{pw_div}
   \mathcal{L}_{PL} = \frac{1}{H \times W}\sum_{k \in \Omega} KL[E(D(e_i^k)) ||T(x_{{aps}_i}^k)]  
\vspace{-5pt}
\end{equation}
where $E(D(e_i^k))$ represents the class probabilities of $k$-th pixel of $i$-th event image, $T(x_{{aps}_i}^k)$ represents the class probabilities of the $k$-th pixel of $i$-th APS image from the teacher $T$, and $\Omega$ = $\{1, 2, . . . , W \times H \}$ denotes all the pixels. 

\begin{figure*}[t!]
\captionsetup{font=small}
    \centering
    \includegraphics[width=.98\textwidth]{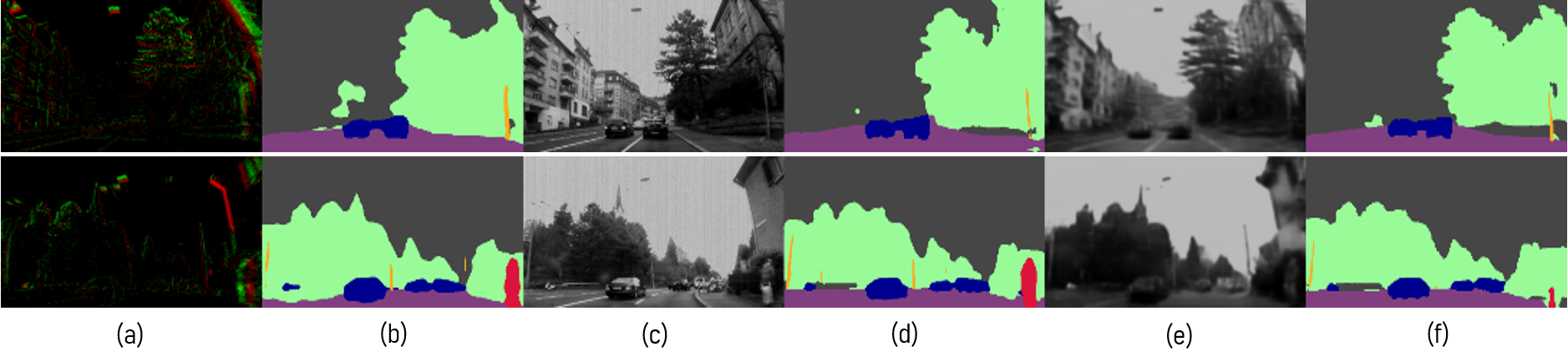}
    \vspace{-7pt}
    \caption{Qualitative results on DDD17 test sequence provided by \cite{alonso2019ev}. (a) Events, (b) Segmentation results on events, (c) APS frames, (d) Segmentation results on APS frames, (e) Generated intensity images from events, (f) Pseudo GT labels.}
\label{fig:ddd17_dataset_evdual}
\vspace{-12pt}
\end{figure*}

\vspace{-6pt}
\subsubsection{Optimization}
The overall objective consists of the supervision loss $\mathcal{L}_{CE}$ in Eq.~\ref{ce_loss} for EEL branch, together with the EIT loss $\mathcal{L}_{EIT}$ in Eq.~\ref{eit_loss} and the SC loss $\mathcal{L}_{SC}$ in Eq.~\ref{sc_loss}. Moreover, it includes the loss terms in the TL module, namely, the feature-level transfer loss $\mathcal{L}_{FL}$ in Eq.~\ref{affinity_graph} and prediction-level transfer loss $\mathcal{L}_{PL}$ in Eq.~\ref{pw_div}. 
The overall objective function is:  
\begin{equation}
\label{total_loss}
    \mathcal{L}=  \mathcal{L}_{CE} + \lambda_1 \mathcal{L}_{EIT} +  \lambda_2 \mathcal{L}_{SC} + \lambda_3 \mathcal{L}_{FL} +  \lambda_4 \mathcal{L}_{PL}
\end{equation}
where $\lambda_1$, $\lambda_2$, $\lambda_3$ and $\lambda_4$ are the hyper-parameters.
We minimize the overall objective with respect to both EEL and EIT branches using dynamic gradient descent strategy.

\section{Experiments and Evaluation}

\vspace{-3pt}
\subsection{Event-based Semantic Segmentation}
\label{seg_exp}
\vspace{-3pt}
Semantic segmentation is the end-task aiming to assign a semantic label, \eg, road, car, in a given scene to each pixel. 

\vspace{2pt}
\noindent \textbf{Datasets.} We use the publicly available driving scene dataset DDD17 \cite{binas2017ddd17}, which includes both events and APS frames recorded by a DAVIS346 event camera. In \cite{alonso2019ev}, 19,840 APS frames are utilized to generate pseudo annotations (6 classes) based on a pretrained network for events (15,950 for training and 3,890 for test).
As the events in the DDD17 dataset are very sparse and noisy, we show more results on the driving sequences in the MVSEC dataset \cite{zhu2018multivehicle}, collected for the 3D scene perception purpose. 

\vspace{2pt}
\noindent \textbf{Implementation details.} 
We use DeepLabv3~\cite{chen2017rethinking} as the semantic segmentation network. The hyper-parameters $\lambda_1$, $\lambda_2$, $\lambda_3$ and $\lambda_4$ are set as 1, 1, 0.1 and 1, respectively. In the training, we set the learning rate as $1e-3$ and use the stochastic gradient descent (SGD) optimizer with weight decay rate of $5e-6$ to avoid overfitting. As the common classification accuracy does not well fit for semantic segmentation, we use the following metric to evaluate the performance, as done in the literature \cite{chen2017deeplab, chen2017rethinking}.  The \textit{intersection of union} (IoU) score is calculated as the ratio of intersection and union between the ground-truth mask and the predicted segmentation mask for each class. We use the \textit{mean IoU} (MIoU) to measure the effectiveness of segmentation.

\vspace{-10pt}
\subsubsection{Evaluation on DDD17 dataset}
\vspace{-3pt}
\noindent \textbf{Comparison.} We first present the experimental results on the DDD17 dataset \cite{alonso2019ev}. We evaluate our method on the test set and vary the window size of events between 10, 50, and 250ms, as done in  \cite{alonso2019ev}. The quantitative and qualitative results are shown in Table~\ref{tab:comp_table1} and Fig.~\ref{fig:ddd17_dataset_evdual}.  We compare our method with two SoTA methods, EvSegNet \cite{alonso2019ev} and Vid2E \cite{gehrig2020video} that uses synthetic version of DDD17 data. Quantitatively, it turns out that the proposed DTL framework significantly improves the segmentation results on events than the baseline (with only CE loss) by around 8\% increase of MIoU. It also surpasses the existing methods by around 4\% increase of MIoU with a multi-channel event representation. 

Meanwhile, on the time interval of 10ms and 250ms, our DTL framework also shows a significant increase of MIoU by around $8.4\%$ and $9.1\%$ than those of the baseline, respectively. The visual results in Fig.~\ref{fig:ddd17_dataset_evdual} further verify the effectiveness of the proposed DTL framework. Overall, the segmentation results on events are comparable to those of the APS frames, and some are even better
\eg, the 1st and 2nd rows. Meanwhile, our method generates convincing intensity images from EIT branch (5th column),
The results indicate that, although events only reflect the edge information,  our method successfully explores the feature-level and prediction-level knowledge to facilitate the end-task learning. The simple yet flexible approach brings a significant performance boost on the end-task learning. 

\vspace{2pt}
\noindent \textbf{High dynamic range (HDR).} HDR is one distinct advantage of an event camera.
We show the segmentation network shows promising performance in the extreme condition.  Figure~4 in the suppl. material shows the qualitative results. The APS frames are over-exposed, and the network fails to segment the urban scenes; however, the events capture the scene details, and the EEL network shows more convincing segmentation results.  

\vspace{2pt}
\noindent \textbf{Segmentation without using GT labels.} With the EIT branch empowered by the teacher model, we show that our DTL framework can learn to segment events without using the semantic labels.  The quantitative and qualitative results are in the suppl. material.  Numerically, even without using the ground truth labels, our method achieves 56.52\% MIoU, which significantly enhances the semantic segmentation performance by 6.60\% MIoU than the baseline. Compared with the SoTA methods \cite{alonso2019ev,gehrig2019video}, our method still surpasses them 
by around 2\% MIoU. 

\begin{table}[t!] 
\renewcommand{\tabcolsep}{8.0pt}
 \centering
 \footnotesize
 \captionsetup{font=small}
 \caption{Segmentation performance of our method and the baseline on the test data \cite{zhu2018multivehicle}, measured by MIoU. The baseline is trained using the pseudo labels made by the APS frames. }
\vspace{-8pt}
 \begin{tabular}{l|c|c}
\hline
 Method & Event representation & MIoU\\
\hline
Baseline-Deeplabv3 & Multi-channel & 50.53 \\
Ours-Deeplabv3 & Multi-channel & \textbf{60.82}  (+ 10.29)\\
\hline
 \end{tabular}
\label{tab:comp_tab_stereo}
\vspace{-14pt}
\end{table}

\begin{figure*}[t!]
\captionsetup{font=small}
    \centering
    \includegraphics[width=\textwidth]{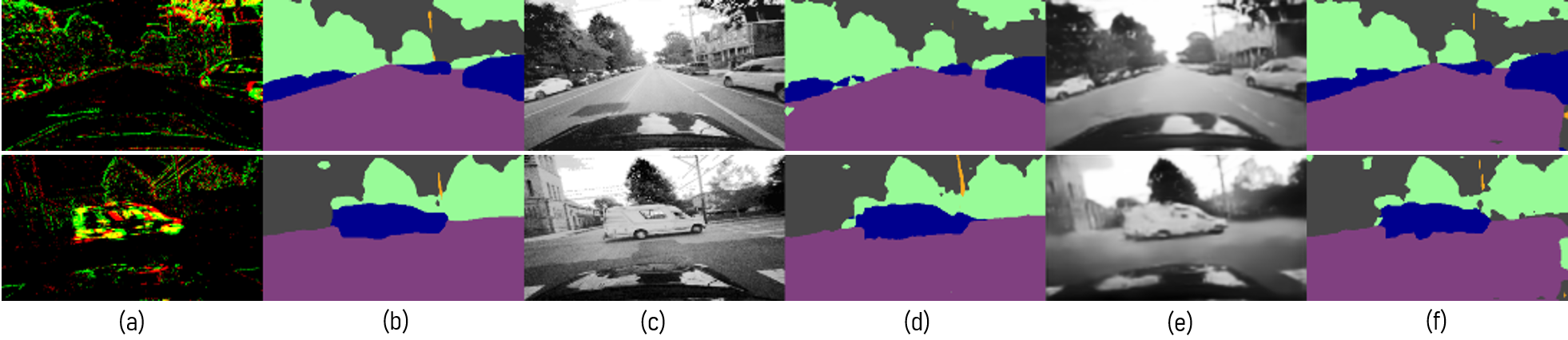}
    \vspace{-22pt}
    \caption{Qualitative results of semantic segmentation and image translation on the MVSEC dataset. (a) Events, (b) Segmentation results on events, (c) APS frames, (d) Segmentation results on APS frames, (e) Generated intensity images from events, (f) Pseudo GT labels.}
    \label{fig:mvsec_results}
    \vspace{-5pt}
\end{figure*}

\begin{table*}[t!] 
\renewcommand{\tabcolsep}{8.0pt}
 \centering
 \footnotesize
 \captionsetup{font=small}
\caption{Quantitative evaluation of monocular dense depth estimation on the MVSEC dataset. }
\vspace{-8pt}
 \begin{tabular}{l|c|c|c|c|c|c|c}
\hline
 \textbf{Method} & \textbf{Dataset} & \textbf{Abs. Rel.} ($\downarrow$) & \textbf{RMSELog} ($\downarrow$)  & \textbf{SILog} ($\downarrow$)  & $\sigma< 1.25$ ($\uparrow$) & $\sigma<1.25^2$ ($\uparrow$) & $\sigma < 1.25^3$ ($\uparrow$) \\
\hline
\cite{zhu2019unsupervised} & \multirow{5}{*}{Outdoor\_day1} & 0.36 & 0.41& 0.16 & 0.46& 0.73 & 0.88\\
\cite{hidalgo2020learning}&  & 0.45&0.63 &0.25&0.47& 0.71 &0.82\\

Translated images & & 0.52 & 0.69 & 0.31 & 0.33& 0.52 & 0.71 \\
Baseline & & 0.33 & 0.39 & 0.16 & 0.63 & 0.80& 0.89 \\
Ours (DTL) & & \textbf{0.29} & \textbf{0.34} & \textbf{0.13} & \textbf{0.71} & \textbf{0.88} & \textbf{0.96} \\
\hline \hline

\cite{zhu2019unsupervised} & \multirow{5}{*}{Outdoor\_night1} & 0.37 & 0.42& 0.15 & 0.45& 0.71 & 0.86 \\
\cite{hidalgo2020learning} & & 0.77&0.64 &0.35&0.33& 0.58 & 0.73 \\
Translated images &  &0.46 & 0.83 & 0.68& 0.26 & 0.49 & 0.69 \\
Baseline &  &0.36 & 0.40 & 0.14& 0.57 & 0.77 & 0.88 \\
Ours (DTL) & & \textbf{0.30}  & \textbf{0.35} & \textbf{0.12} & \textbf{0.69} & \textbf{0.88} & \textbf{0.95} \\
\hline
\end{tabular}
\label{tab:comp_tab_depth}
\vspace{-10pt}
\end{table*}

\vspace{-10pt}
\subsubsection{Evaluation on MVSEC dataset}
\vspace{-3pt}
To further validate the effectiveness of the proposed DTL framework, we show more results on the MVSEC dataset \cite{zhu2018multivehicle}.
As there are no segmentation labels in this dataset, to numerically evaluate our method, we utilize the APS frames to generate pseudo labels based on a network \cite{chen2017rethinking}, 
similar to \cite{alonso2019ev}, as our comparison baseline. Due to the poor quality of APS frames in the outdoor\_day1 sequence, we mainly use outdoor\_day2 sequence and divide the data into training (around 10K paired event images and APS frames) and test (378 paired event images and APS frames) sets based on the way of splitting DDD17 dataset in \cite{alonso2019ev}. For the training data, we remove the redundant sequences, such as vehicles stopping in the traffic lights, etc.  We also use the night driving sequences to show the advantage of events on HDR. 
 
The qualitative and quantitative results are shown in Fig.~\ref{fig:mvsec_results} and Table~\ref{tab:comp_tab_stereo}. In Fig.~\ref{fig:mvsec_results}, we mainly show the results in the general condition. The experimental results of HDR scenes are provided in suppl. material. 
Using a multi-channel event representation in Table~\ref{tab:comp_tab_stereo}, the proposed DTL framework significantly surpasses the baseline by a noticeable margin with around 10.3\% increase of MIoU. The results indicate a significant performance boost for semantic segmentation. The effectiveness can also be verified from visual results in Fig.~\ref{fig:mvsec_results}. As can be seen, the semantic segmentation results (2nd column) are fairly convincing compared with the results on APS frames (4th column) and the pseudo GT labels (6th column). Meanwhile, our method also generates very realistic intensity images (5th column) from the EIT branch.
The results on both semantic segmentation and image translation show that our the proposed DTL framework successfully exploit the knowledge from one branch to enhance the performance of the other.

\vspace{-3pt}
\subsection{Monocular Dense Depth Estimation}
\vspace{-2pt}
\label{depth_sec}
Depth estimation is the end-task of predicting the depth of scene at each pixel in the image plane.
Previous works for event-based depth estimation have most focused on sparse or semi-dense depth estimation \cite{rebecq2016emvs,zhou2018semi,rebecq2016evo,rebecq2017real}. Recently, DNNs have been applied to stereo events to generate dense depth predictions \cite{tulyakov2019learning} and to estimate monocular semi-dense depth \cite{zhu2019unsupervised}. Some other works have focused on the dense depth estimation with only events \cite{hidalgo2020learning} or with additional inputs \cite{gehrig2021combining}. 
We show that the proposed DTL framework is capable of predicting monocular dense depth from sparse event data. To evaluate the scale-invariant depth, we use the absolute relative error (Abs. Rel.), logarithmic mean squared error (RMSELog), scale invariant logarithmic error (SILog) and accuracy (Acc.).

We present quantitative and qualitative results and compare with the baseline settings and prior methods \cite{zhu2019unsupervised,hidalgo2020learning} with sparse event data as inputs on the MVSEC dataset \cite{zhu2018multivehicle}.
We use outdoor\_day2 sequence of the MVSEC dataset where we select around 10K embedded event image and APS image pairs with their synchronized depth GT images to train the our DTL framework, similar to \cite{zhu2019unsupervised}. 
We then utilize the outdoor\_day1 sequence (normal driving condition) and night driving sequences as the test sets. More details about dataset preparation and implementation (\eg, network structure, loss functions) are in the suppl. material.  

\begin{figure*}[t!]
\captionsetup{font=small}
\centering
\includegraphics[width=.96\textwidth]{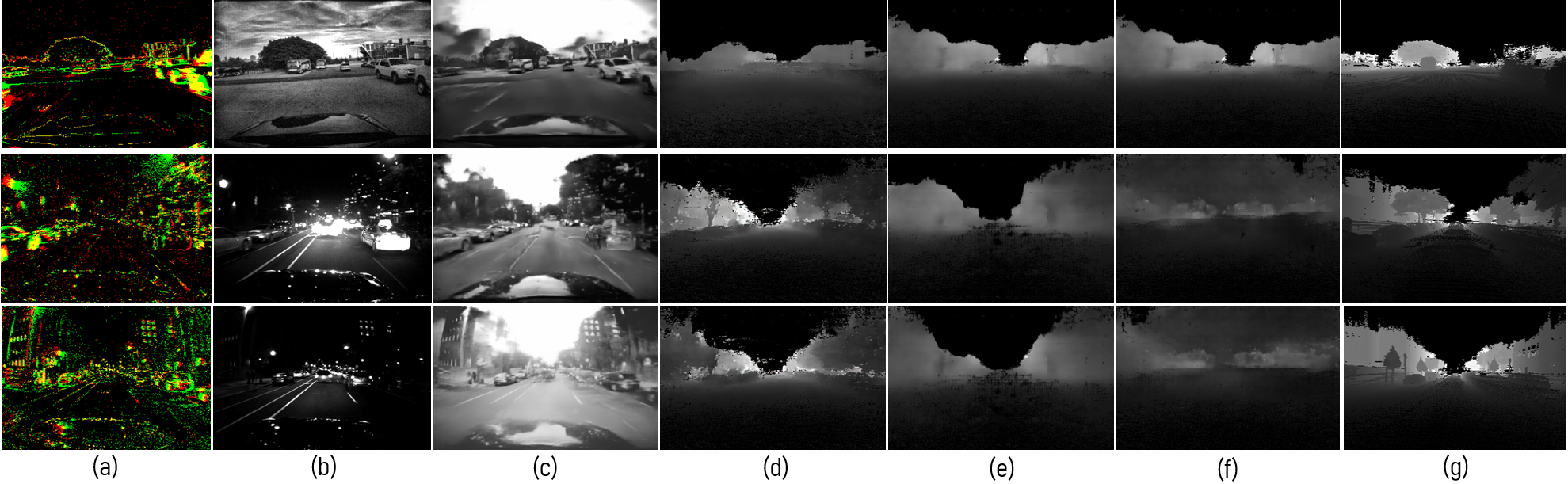}
\vspace{-8pt}
\caption{Qualitative results for monocular dense depth estimation. (a) Events, (b) Dark APS frames, (c) Generated intensity images, (d) Predicted depth on events, (e) Predicted depth on the generated intensity images, (f) Predicted depth on APS frames, (g)  GT depth.}
\label{fig:depth_res}
\vspace{-10pt}
\end{figure*}

Table~\ref{tab:comp_tab_depth} shows the quantitative results, which are supported by the qualitative results in Fig.~\ref{fig:depth_res}. On the outdoor\_day1 sequence, our method achieves around 10\% Abs. Rel. drop than the baseline and the compared methods.
The effectiveness on outdoor\_day1 sequence  can also be verified from Fig.~\ref{fig:depth_res} (1st row). 
Compared with the GT depth, 
our method predicts depth with clear edges and better preserves the shapes and structures of objects, such as buildings, trees, cars, etc.  Meanwhile, our method is also capable of translating events to high-quality intensity images, where we can see the translated images are close to the APS frames.

\vspace{2pt}
\noindent \textbf{HDR depth.} Our method shows apparent advantages on the HDR scene. As shown in Fig.~\ref{fig:depth_res} (2nd and 3rd rows), when the APS frames fail to predict the correct depth information (6th column), events show promising depth estimation results (4th columns). Moreover, our method generates realistic intensity images (3rd column) and shows better depth information (5th column) than those of APS frames. In particular, when the APS frames are almost invisible, our method generates convincing HDR  images that better preserve the structures of objects, such as buildings, trees, cars, etc. The effectiveness can also be numerically verified in Table~\ref{tab:comp_tab_depth}. Our method achieves around 17\% performance boost (\eg, Abs. Rel.) than those of SoTA methods~\cite{zhu2019unsupervised,hidalgo2020learning}, the baseline and generated intensity images.

\vspace{-4pt}
\section{Ablation Study and Analyses}
\vspace{-3pt}
\noindent \textbf{Loss functions.}
We first study the effectiveness of adding and removing the loss terms in Eq.~\ref{total_loss}. For convenience, we mainly focus on semantic segmentation on the DDD17 dataset. The ablation results are shown in Table~\ref{tab:ablation_table}. In general, without TL module, the EIT branch slightly improves the EEL branch. However, with feature-level transfer loss in the TL module, the performance of segmentation is significantly enhanced by around 5.58\% increase of MIoU. When the predication-level transfer loss is added, EEL performance is further enhanced to MIoU of 58.80\%. From the ablation study, it clearly shows that our DTL framework is a successful approach for benefiting the end-task learning.

\begin{figure}[t!]
\captionsetup{font=small}
    \centering
    \includegraphics[width=.97\columnwidth]{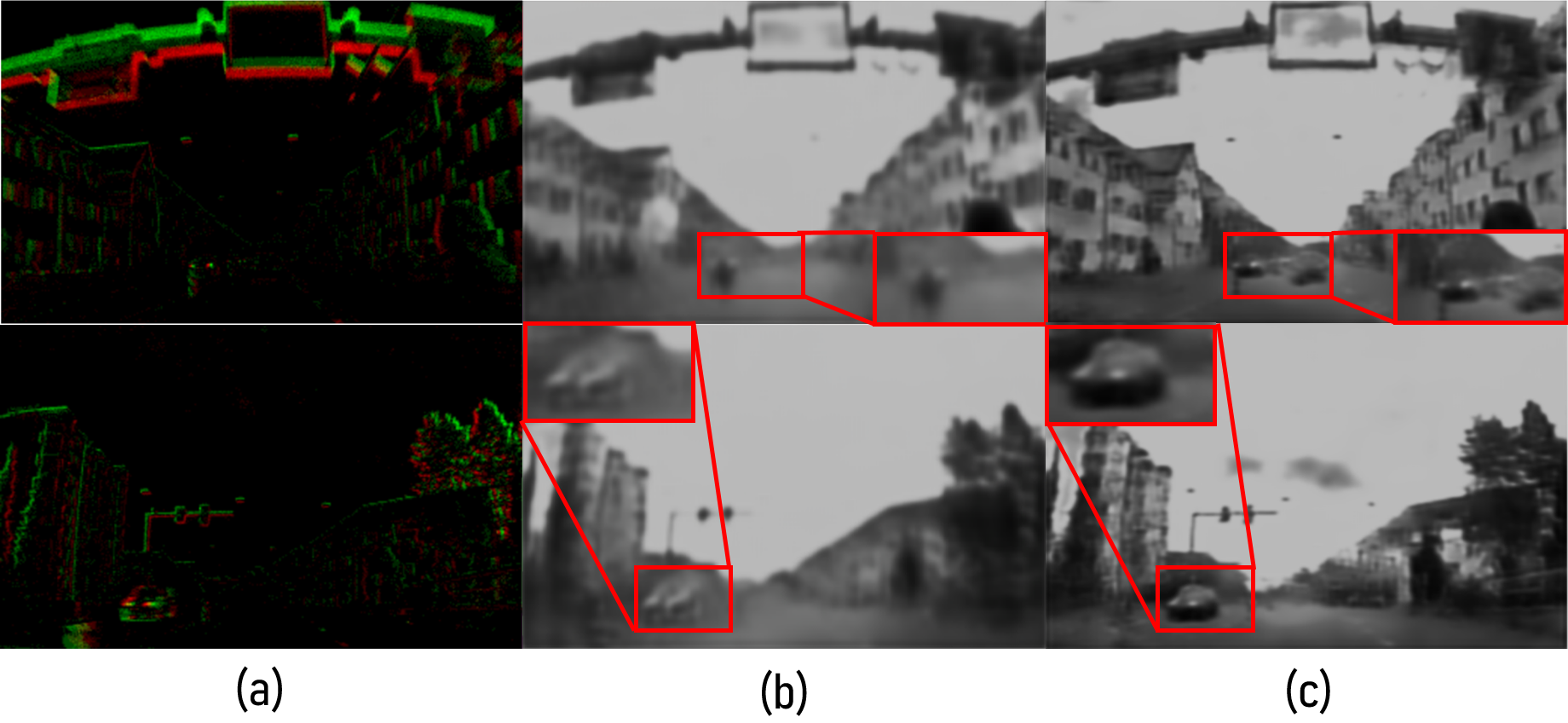}
    \vspace{-10pt}
    \caption{Impact of DTL on image translation. (a) Events, (b) Translated images without DTL, (c) Translated images with DTL.}
    \label{fig:eit_ablation}
    \vspace{-5pt}
\end{figure}

\vspace{2pt}
\noindent \textbf{The effectiveness of TL module for EIT branch.}
Although the EIT branch is regarded as an \textit{auxiliary} task in the proposed DTL framework, we show that it also benefits the EIT learning. We qualitatively compare the quality of translated images with and without using the DTL framework. Fig.~\ref{fig:eit_ablation} shows the visual results. In contrast to the generated images without DTL (2nd column), the results with DTL are shown to have more complete semantic information and better structural details, as shown in the cropped patches in the 3rd column. Interestingly, better structural details, \eg, cars, buildings and trees, are restored. The experimental results show that our method works effectively on sparse events and are shown successful not only for the end-tasks but also for the image translation.

\vspace{2pt}
\noindent \textbf{Event representation vs. EEL performance.}
We now study how event representation impacts the performance on end-task learning under the DTL framework. We leverage several existing event representation methods, \eg, voxel grid \cite{zhu2018unsupervised} and 6-channel \cite{alonso2019ev} and the multi-channel event embedding methods \cite{wang2020event,wang2019event} used in the paper. For convenience, we compare the efficacy of these methods on semantic segmentation task using the DDD17 dataset. The numerical results are shown in Table~\ref{tab:event_rep_vs_perf}. In general, event embedding methods have a considerable influence on semantic segmentation performance. Overall, the multi-channel representation is demonstrated to show sightly better segmentation performance than the other two methods. 

\begin{table}[t!] 
\renewcommand{\tabcolsep}{25.0pt}
 \centering
 \footnotesize
 \captionsetup{font=small}
 \caption{Ablation study results of the proposed DTL framework based on DDD17 dataset. }
\vspace{-8pt}
 \begin{tabular}{l|c}
\hline
 Module & MIoU [50ms]\\
\hline
CE & 50.92 \\
CE + EIT & 53.87 (+2.95)\\
CE + EIT + CS & 54.66 (+3.74) \\
CE + EIT + TL (FL) & 56.50 (+5.58)\\
CE + EIT + TL (FL + PL) &\textbf{58.80} (+7.88)\\
\hline
 \end{tabular}
\label{tab:ablation_table}
\vspace{-5pt}
\end{table}

\begin{table}[t!] 
\renewcommand{\tabcolsep}{10.0pt}
 \centering
 \footnotesize
 \captionsetup{font=small}
 \caption{The impact of event representations on the semantic segmentation performance on DDD17 dataset. }
\vspace{-8pt}
 \begin{tabular}{l|c|c}
\hline
 Method & Event Rep. & MIoU [50ms]\\
\hline
Ours-Deeplabv3 & Voxel grid \cite{zhu2018unsupervised} & 56.30\\
Ours-Deeplabv3 & 6-channel \cite{alonso2019ev} &57.68\\
Ours-Deeplabv3 & Multi-channel & \textbf{58.80}\\
\hline
 \end{tabular}
\label{tab:event_rep_vs_perf}
\vspace{-14pt}
\end{table}

\begin{figure}[t!]
\captionsetup{font=small}
    \centering
    \includegraphics[width=\columnwidth]{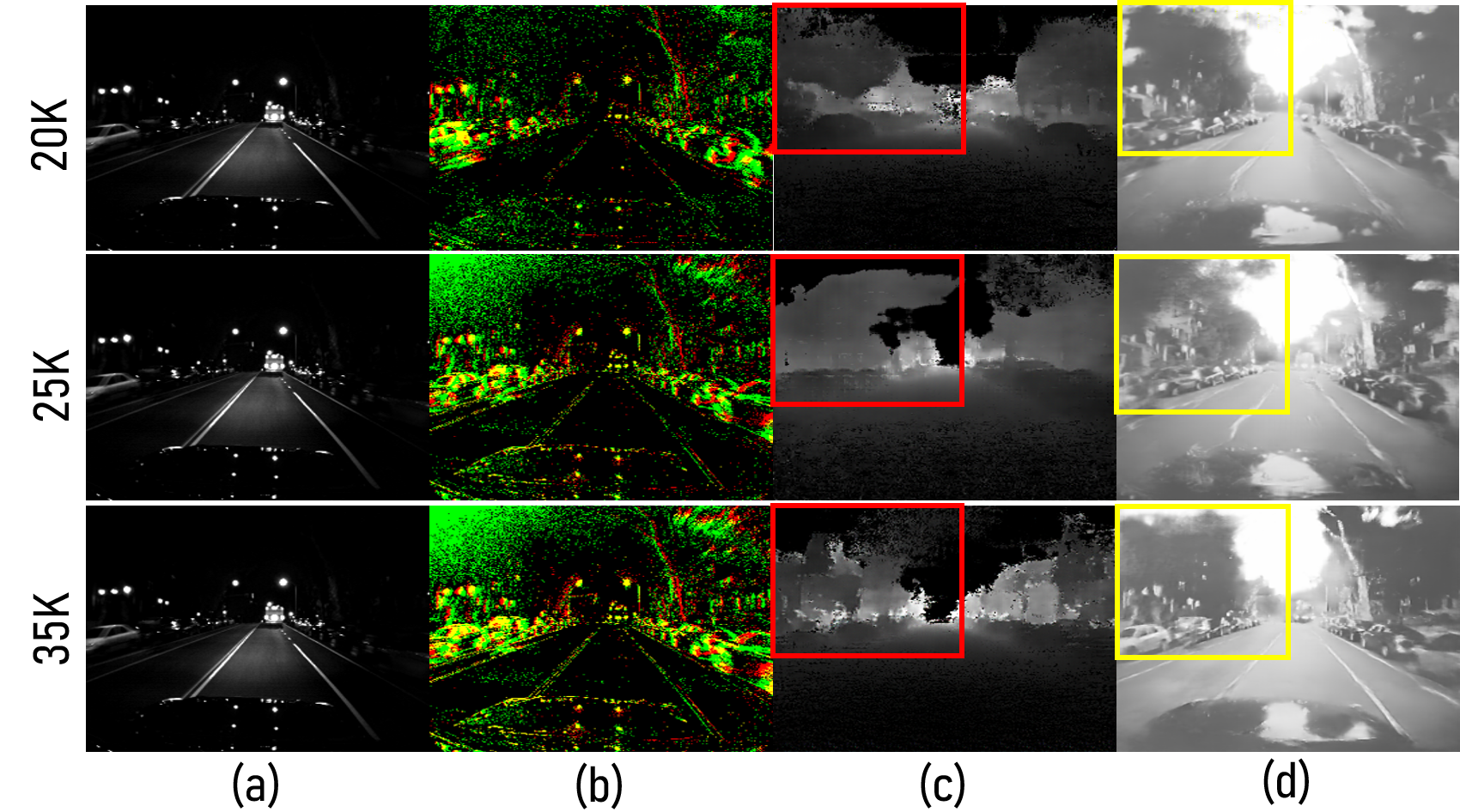}
    \vspace{-20pt}
    \caption{Impact of number of events (top: 20K; middle: 25K; bottom: 35K) on our DTL framework. (a) APS frames, (b) Events, (c) Predicted depth on events, (d) Generated intensity images. }
    \label{fig:num_performace}
    \vspace{-15pt}
\end{figure}
\vspace{1pt}
\noindent \textbf{Number of events vs. overall performance.}
The number of events used for event representation also impacts the performance on the end-task learning and image translation. We thus conduct an analysis on how the number of events used for event representation affects the end-task, \eg, dense depth estimation. Fig.~\ref{fig:num_performace} shows the visual comparisons on the  outdoor\_night2 sequence of the MVSEC dataset, in which we highlight the HDR capability. 
In particular, the results of embedding 35K events of the bottom row show better depth estimation (as shown in the red box) and intensity image translation (as shown in the yellow box) results than those of 25K and 20K events, respectively. 

\vspace{-3pt}
\section{Conclusion and Future work}
\vspace{-3pt}
In this paper, we presented a simple yet novel two-stream framework, named DTL for promoting end-task learning, with no extra inference cost. The DTL framework consists of three components: the EEL, the EIT and TL module, which enriches the feature-level and prediction-level knowledge from EIT to improve the EEL. The simple method leads to strong representations of events and is evidenced by the promising performance on two typical tasks. 
As the DTL framework is a general approach, we plan to apply it to other modality data, such as depth and thermal data.

\noindent \textbf{Acknowledgement.} \small{This work was supported by the National Research Foundation of Korea(NRF) grant funded by the Korea government(MSIT) (NRF-2018R1A2B3008640)}.

{\small
\bibliographystyle{ieee_fullname}
\bibliography{egbib}
}

\end{document}